\documentclass[conference]{IEEEtran}
\IEEEoverridecommandlockouts
\usepackage{cite}

\usepackage{graphicx}
\usepackage{textcomp}
\usepackage{booktabs}
\usepackage{color}
\usepackage{float}
\usepackage{multirow}
\usepackage{subcaption}
\usepackage{caption}

\usepackage{fancyhdr}
\def\BibTeX{{\rm B\kern-.05em{\sc i\kern-.025em b}\kern-.08em
    T\kern-.1667em\lower.7ex\hbox{E}\kern-.125emX}}

\begin{document}

\title{AdaCon: Adaptive Context-Aware Object Detection for Resource-Constrained Embedded Devices}

\author{\IEEEauthorblockN{Marina Neseem}
\IEEEauthorblockA{\textit{School of Engineering} \\
\textit{Brown University}\\
Providence, RI \\
marina\_neseem@brown.edu} \\
\and
\IEEEauthorblockN{Sherief Reda}
\IEEEauthorblockA{\textit{School of Engineering} \\
\textit{Brown University}\\
Providence, RI \\
sherief\_reda@brown.edu} \\}

\IEEEaftertitletext{\vspace{-3.2\baselineskip}}

\maketitle

\thispagestyle{fancy}
\cfoot{\thepage}

\begin{abstract}
Convolutional Neural Networks achieve state-of-the-art accuracy in
object detection tasks. However, they have large computational and energy requirements that challenge their deployment on resource-constrained edge devices. Object detection takes an image as an input, and identifies the  existing object classes as well as their locations in the image. In this paper, we leverage the prior knowledge about the probabilities that different object categories can occur jointly to increase the efficiency of object detection models. In particular, our technique clusters the object categories based on their spatial co-occurrence probability. We use those clusters to design an adaptive network. During runtime, a branch controller decides which part(s) of the network to execute based on the spatial context of the input frame. Our experiments using COCO dataset show that our adaptive object detection model achieves up to 45\% reduction in the energy consumption, and up to 27\% reduction in the latency, with a small loss in the average precision (AP) of object detection.

\end{abstract}


\section{\textbf{Introduction}}
\label{sec:intro}
Convolutional neural networks (CNNs) are used to achieve high accuracy in object detection tasks \cite{redmon2018yolov3,ren2015faster}.
In object detection, the input is an image, and the outputs are the object classes and their location in the image represented by the bounding box coordinates.
CNN-based object detectors can be divided into two main categories: two-stage detectors \cite{ren2015faster,he2017mask} and one-stage detectors \cite{redmon2018yolov3,retiannet}.
Two-stage detectors first identify some regions of interest (ROIs) as candidates, and then classify and regress those ROIs to identify the objects and their bounding boxes.
One-stage detectors directly predict the object categories and the bounding boxes using some default anchors.
One-stage detectors are less accurate than two-stage detectors, but they are more efficient; hence they are more suitable for edge devices.

In recent years, the deployment of CNNs on edge devices like mobile phones, smart glasses, and augmented reality devices has become essential for real-time response and for data privacy reasons.
However, those networks have high memory and computational power demands, which challenge their deployment on resource-constrained embedded devices.
These devices are battery-operated, so low energy consumption is crucial.
They also have memory constraints, so the model should be compact.
Finally, those models have real-time constraints when deployed on edge devices.

To reduce the memory and computational demands of CNNs, researchers have explored pruning and quantization of those models to create a trade-off among energy, latency, and accuracy \cite{pruned_yolo}.
Others use dynamic networks \cite{chhay-configurable, yu2018slimmable, zhang2019domain, Tann2018FlexibleDN} that can be reconfigured during runtime to accommodate the target accuracy and efficiency.
Another effective approach is to use hierarchical networks \cite{treecnn}.
Those networks have shown success in image classification tasks because each image has only one object, so the objects can be grouped based on their visual or semantic similarities \cite{goel2020low}.
However, for object detection, an image can have objects that are not visually or semantically similar.
Therefore, the previously proposed methods would not work for object detection. 
For example, a scene with a parking meter next to a car is very likely to happen.
None of the techniques based on visual or semantic similarity will group the car and the parking meter together.
Another example is a scene with a couch and a television.
However, in a spatial-context-based approach, a car and a parking meter will be grouped together, while a couch and a television will form another group.
Thus, we propose using the spatial context to group the object classes, this would allow our adaptive model to efficiently complete the detection task by executing a single branch, making it appropriate for resource-constrained devices.

In this paper, we propose a novel approach that leverages the information about the spatial context of the objects to design an efficient adaptive model for object detection. 
Our adaptive model reduces the energy consumption, the latency, and the memory footprint for object detection with a negligible loss in accuracy.
Our contributions can be summarized as follows:
\begin{itemize}
    \item To the best of our knowledge, AdaCon is the first work to introduce an adaptive methodology for one-stage object detectors handling images with multiple objects. Our adaptive method enables efficient object detection on resource-constrained embedded devices.
    \item AdaCon leverages the information about the spatial-context of the object categories to construct a knowledge graph. Then, we use the constructed knowledge graph to design our adaptive context-aware object detection model.
    \item We introduce a simple yet effective and generalizable methodology that can be easily applied to design the neural network architecture of the detection branches for our adaptive model.
    \item We apply our adaptive methodology to two different state-of-the-art object detectors, and we compare different generated adaptive models representing different energy-accuracy trade-offs to the static baselines. 
    \item We deploy our AdaCon models on an embedded Nvidia Jetson nano board, and analyze the accuracy, latency, and energy of the different architectures. Our adaptive model achieves up to 45\% reduction in energy, and up to 27\% reduction in the latency with small loss in average precision on the COCO dataset. 
\end{itemize}

The rest of the paper is organized as follows. We review the related work in Section~\ref{sec:previous_work}. Then, we introduce our adaptive object detection technique in Section~\ref{sec:clustering}. Next, we show the experimental setup and results in Section~\ref{sec:experiments}. Finally, we conclude in Section~\ref{sec:conclusion}.

\begin{figure*}[t]
\centerline{\includegraphics[width=0.85\textwidth, keepaspectratio]{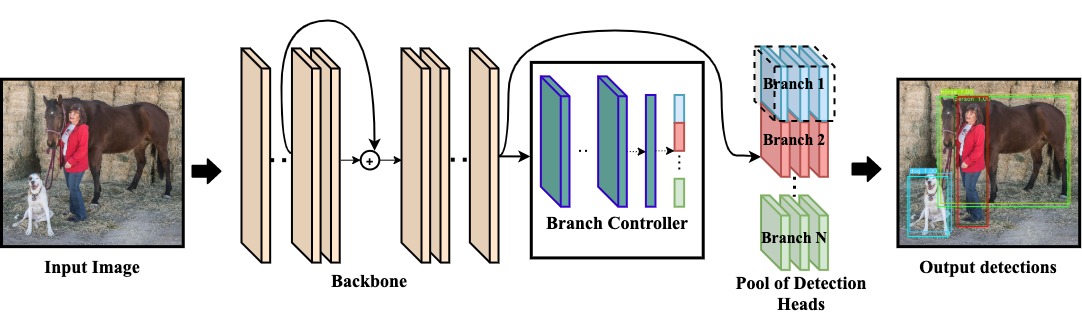}}
\caption{Illustration of our Adaptive Object Detection model. The backbone is first executed to extract features from the input image. Then, the branch controller takes the extracted features, and route them towards one or more of the downstream detection heads. Only the chosen head(s) are then executed to get the detected object categories and their bounding boxes.}
\label{adaptive_detection}
\vspace{-10pt}
\end{figure*}

\section{\textbf{Previous Work}}
\label{sec:previous_work}

\noindent \textbf{Object Detection:} Object detection is a well-researched topic because it is essential in many applications like augmented reality, surveillance, and autonomous driving. 
However, most object detectors are not suitable for embedded or wearable devices because they are based on complex power-hungry DNNs with large memory footprint. 
One-stage detectors such as YOLOv3 \cite{redmon2018yolov3} and RetinaNet \cite{retiannet} are faster than most two-stage detectors like Faster R-CNN \cite{ren2015faster} and Mask R-CNN \cite{he2017mask}, so they are more suitable for real-time response on edge devices. 
Thus, we choose YOLOv3 \cite{redmon2018yolov3}, and RetinaNet \cite{retiannet} as a the baseline architecture on which we apply our method for adaptive context-aware object detection.

\noindent \textbf{Efficient Neural Networks:} Over the past few years, the need for efficient computing on edge devices increased, so researchers started designing compact networks \cite{howard2017mobilenets,ma2018shufflenet, wu2017squeezedet}. Others used pruning and quantization \cite{fedorov2019sparse, pruned_yolo}. In this paper, we consider a different research direction for optimization by exploring adaptive networks that can leverage the nature of the object detection task. Our approach is an orthogonal effort to build more efficient neural networks that are suitable for constrained embedded devices.

\noindent \textbf{Dynamic/Adaptive Neural Networks:} Adaptive neural networks have been adopted by many researchers to reduce the computational complexity needed for neural networks, hence reduce the latency and energy requirements without sacrificing much accuracy. Tann \emph{et al.} \cite{chhay-configurable, Tann2018FlexibleDN} and Yu \emph{et al.} \cite{yu2018slimmable} trained a single network at different widths to permit adaptive accuracy-energy trade-offs at runtime. Others skip some layers \cite{wang2018skipnet}, or deploy some early exit criteria \cite{bolukbasi2017adaptive} to reduce the computation complexity. 
Recently, Zhang \emph{et al} \cite{zhang2019domain} proposed building Domain-Aware networks that decide which part of the network to run based on the weather and the time of the day in which the device is operating.
All these methods are generic for convolutional neural networks and image classification, but they have barely been applied to object detection. 
Our work is different because we leverage the information about the spatial context of the object categories.
The spatial context implies the probability that different objects can occur jointly, and this is an essential information that can be used to further optimize object detection models.

\noindent \textbf{Context-aware Object Detection:}
On another research thread, some researchers have been exploring the use of prior knowledge about the real world to improve the accuracy of object detection models. Fang \emph{et al} \cite{fang2017object}, and Xu \emph{et al} \cite{xu2019spatial} integrate knowledge-graphs and adjacency-matrices to leverage the information about the co-occurrence and the locations of the objects for more accurate object detection. Their main goal is to increase the model accuracy, and they achieve this by adding more computational complexity, which further reduces the model efficiency. However, our criteria is different, because we exploit the idea of using the prior knowledge to build efficient adaptive object detection models. Our adaptive model architecture reduces the energy consumption, latency, and runtime memory requirements of object detectors, making them more appropriate for resource-constrained devices.

\section{\textbf{Proposed Adaptive Context-aware Object Detection}}
\label{sec:clustering}
Modern object detectors are typically composed of two main parts: a \emph{backbone} which extracts the features from the input image, and a \emph{head} which is used to predict the object categories and their bounding boxes. 
In this work, we propose an adaptive context-aware neural network architecture for object detection. 
We achieve this architecture by leveraging the information about the co-occurrence of objects in the spatial domain while designing our object detection model. 

Our adaptive model consists of two main components: spatial-context-based clustering and a hierarchical object detection model.
In the spatial-context-based clustering, we extract the information about the co-occurrence of object categories from the training data, and use this information to cluster the object categories, where each cluster has the objects that co-occur in the spatial domain with high probability.
Then, we leverage the extracted information to design an adaptive object detection model.

As illustrated in Figure \ref{adaptive_detection}, our adaptive object detection model consists of three main components: \emph{backbone}, \emph{branch controller}, and a pool of\emph{ specialized branches}.
The backbone is a CNN responsible for extracting the features from the input image. 
The branch controller is a regression model with a sigmoid activation function at its output.
The branch controller has an input size similar to the output feature maps of the backbone, and the number of its outputs is equal to the number of clusters chosen while running the spatial-context-based clustering. 
Those outputs represent the \emph{confidence score} that the input image belongs to the corresponding spatial context. 
Using the confidence scores in the branch controller, we enable two modes of operation: \emph{single-branch execution} where only the branch with the highest score is selected, and \emph{multi-branch execution} where all the branches with scores higher than a certain threshold are selected. 
The final component of our model is a pool of specialized detection heads (branches).
Each specialized branch is responsible for detecting the objects that belong to one spatial context. 

During run-time, the input image first passes through the backbone, then the output of the backbone is passed to the branch controller which classifies those feature maps to one or more spatial contexts. 
Receiving the decision from the branch controller, we execute the chosen branch(es), and concatenate their outputs to complete the detection task.
In this Section, we are going to explain in detail the clustering method in Section \ref{context_based_clustering}. 
Section \ref{branch_architecture} shows our method for selecting the architecture of the specialized branches. 
Finally, we explain the training method for our adaptive object detection model in Section \ref{training}.

\begin{figure*}[t]
\abovecaptionskip -5 pt
\centerline{\includegraphics[clip,height=100pt,keepaspectratio]{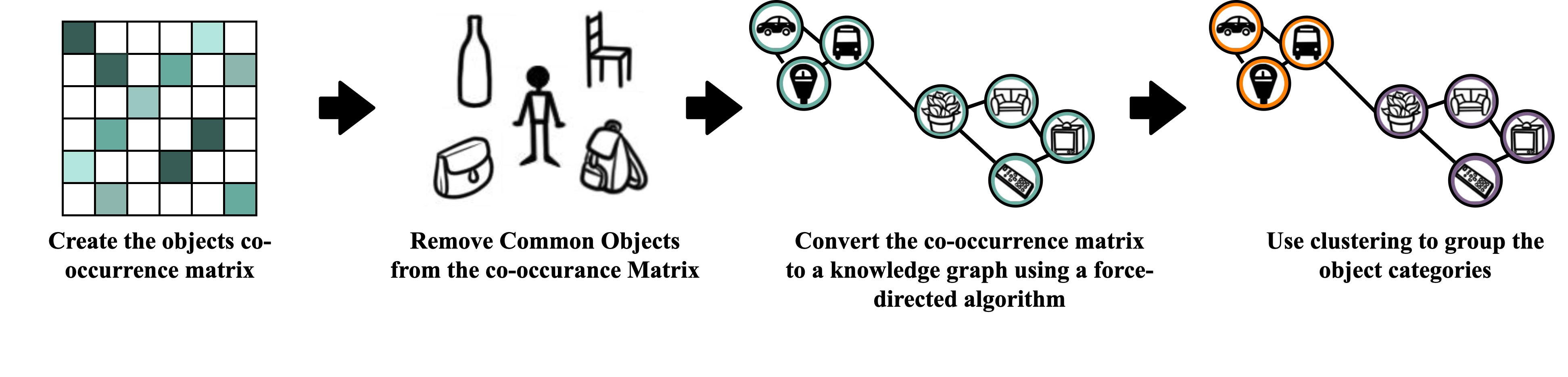}}
\caption{Spatial-Context-based clustering for the object categories.}
\label{clustering_technique}
\end{figure*}

\vspace{-5pt}
\subsection{\textbf{Spatial-Context based Clustering:}}
\label{context_based_clustering}
The ultimate goal of our clustering technique is to group objects that can co-occur in the same scene.
Then, we use those clusters to construct our hierarchical detection model where each cluster has a corresponding branch in the model.
This would guarantee high performance and high efficiency because it would enable detecting most of the objects in a scene by executing only one branch of the detection model. 
Figure \ref{clustering_technique} shows our spatial-context-based clustering technique. First, we construct the co-occurrence matrix of the object categories where each value represents the frequency of the co-occurrence of the object categories in the same scene across all the training dataset.
Then, we extract the common objects. 
Those are the object categories that have high probability of co-occurrence with more than 75\% of other object categories. 
We remove those common objects from the co-occurrence matrix, and add them later to each cluster. Our intuition is that some objects such as a ``person'' or a ``backpack'' can co-occur with any other object in any spatial setup. 
Thus, it is better that our network considers the presence of those common objects all the time. 
Next, we convert the frequency-based co-occurrence matrix to a correlation matrix. Then, we use the correlation matrix to build a knowledge graph using Fruchterman-Reingold force-directed algorithm \cite{fruchterman1991graph}. 
As shown in Figure \ref{clustering_technique}, the nodes of the knowledge graph represent the object classes, and the edges represent the probability of the co-occurrence of the connected objects in the same scene. 
Afterwards, we use agglomerative clustering to group the nodes based on their location in the knowledge graph. 
This results in clusters of object classes where the inter-cluster objects have low probability of appearing together in the same scene, while the intra-cluster objects have a high likelihood of occurring jointly.

\subsection{\textbf{Detection Head Architecture}}
\label{branch_architecture}
In our hierarchical adaptive model, each branch (i.e., detection head) is responsible for detecting a subset of the object classes. 
The number of object classes assigned to a certain branch depends on the total number of branches in the adaptive model, as well as the outcome of the spatial-context clustering as explained in Section \ref{context_based_clustering}.
This means that for a given model with a certain number of branches, the object classes are not equally distributed among those branches.
Therefore, the number of object classes should be considered while designing the branch architecture to guarantee that each branch has an appropriate representational capacity to make it efficient without sacrificing accuracy.

We propose a systematic approach to choose the detection head architecture. Our approach can be easily generalized to different object detection models.
For a given model, we choose the head architecture of the static baseline model as our template.
For each branch detection head, we define \textit{a compression factor} equal to the number of object classes assigned to this branch, divided by the total number of object classes. 
Then for each layer in the template head architecture, we keep the same number of layers, but we compress the model by reducing the number of channels in each layer by the previously calculated compression factor.

For example, assuming that we use a two-branch AdaCon model to detect 30 object classes, and our spatial-context-based clustering assigned 18 and 12 objects to the first and the second branches, respectively.
Then, the compression factor for the first branch is $18/30$, while it is $12/30$ for the second branch. 
Table \ref{branch_arch_search} compares the number of parameters, and the number of multiply-and-accumulate (MAC) operations for the compressed branches, as well as the static baseline template for YOLOv3 \cite{redmon2018yolov3} and RetinaNet \cite{retiannet}. 
We can notice that the total number of parameters and MAC operations for the branches are always smaller than the corresponding template. 
Moreover, the various branches would not be active at the same time.
Therefore, the dynamic number of parameters and MAC operations at a time would be significantly smaller, which reduces the memory footprint as well as the latency and the energy. More analysis is presented in Section \ref{sec:experiments}. 

\begin{center}

\begin{table}[h]
\caption{Detection Head (Branches) Architectures for AdaCon models with various number of branches.}
\centering
\renewcommand{\arraystretch}{1.1}
\setlength{\tabcolsep}{2.5pt}
\begin{tabular}{l l | l | l} 
\toprule
 \multicolumn{2}{c}{\textbf{Model}} & \textbf{Branches Parameters} & \textbf{Branches MACs} \\
 & & \textbf{$(M)$} & \textbf{$(G ops)$} \\
\midrule
\midrule

\multirow{5}{*}{RetinaNet} & Baseline & 6.7 & 22.5 \\
                        & $2$  Branches &  3.0, 2.1 & 4.2, 2.9 \\
                        & $3$  Branches & 2.1, 1.4, 1.4 & 2.9, 1.9, 1.9 \\ 
                        & $4$  Branches & 1.7, 1.4, 1.4, 0.5 & 2.4, 1.9, 1.9, 0.7 \\ 
                        & $5$  Branches & 1.4, 1.3, 1.4, 0.5, 0.6 & 1.9, 1.8, 1.9, 0.7, 0.9 \\ 
\midrule
\multirow{5}{*}{YOLOv3} & Baseline & 21.3  & 8.4 \\
                        & $2$  Branches & 6.0, 9.2 & 2.4, 3.6 \\
                        & $3$  Branches & 6.0, 3.8, 3.8 & 2.4, 1.5, 1.5 \\ 
                        & $4$  Branches & 5.0, 3.8, 3.8, 1.2 & 1.9, 1.5, 1.5, 0.5 \\ 
                        & $5$  Branches & 3.8, 3.6, 3.8, 1.2, 1.5 & 1.5, 1.4, 1.5, 0.5, 0.6 \\ 
\bottomrule
\end{tabular}
\label{branch_arch_search}
\vspace{-15pt}
\end{table}
\end{center}

\vspace{-25pt}

\subsection{\textbf{Training the Adaptive Object Detection Model}}
\label{training}
Each module of our adaptive network is trained separately with the correct pairs of inputs and labels depending on the task assigned to this module.
We use a multi-stage training technique to train our adaptive model. 
In stage 1, we train the backbone. 
Then in stage 2, we freeze the backbone, and concurrently train our branch controller as well as the detection branches.

\noindent \textbf{Training the backbone:} In our implementation, we train the backbone as a part of the static model. 
We then take the pre-trained backbone and use it as the backbone of our adaptive model. 

\noindent \textbf{Training the branch controller:} The branch controller is a regression model with few convolutional layers and a sigmoid activation layer at the output. 
It predicts the probability that the input image belongs to each spatial context. 
As shown in Figure \ref{adaptive_detection}, the inputs to the branch controller are the feature maps generated by the backbone for each image, and the labels are the dominant spatial-context on each image.
To generate the labels for training the branch controller, we map the objects in every image to their spatial-context according to the output of the spatial-context-based clustering, and we label the image with the dominant spatial context.

\noindent \textbf{Training the detection branches:} The detection branches are trained concurrently on the relevant object categories according to the output of the spatial-context-based clustering proposed in Section \ref{context_based_clustering}.
The input to each branch is also the feature maps generated by the backbone for each image, and the labels are the bounding boxes and the object categories in the image.

\section{\textbf{Experimental Evaluation and Results}}
\label{sec:experiments}
\subsection{\textbf{Experimental Setup}}

\noindent \textbf{Dataset:} We evaluate our method using the Microsoft COCO dataset \cite{lin2014microsoft}. 
This dataset has over 120K training images with 80 different object categories. The 80 categories cover a wide range of indoor (i.e., bedroom, kitchen, bathroom, living room, office, etc.) and outdoor scenes (i.e., street, park, farm, zoo, etc.).
The images in the COCO dataset are all collected from real world scenes with some occlusions, and under various lighting conditions.
Those images are annotated with the bounding boxes for the different object categories.

\noindent \textbf{Implementation:} We implemented our proposed technique using PyTorch.
We used our spatial-context-based clustering to cluster the 80 objects of COCO dataset into different number of clusters as mentioned in Section \ref{context_based_clustering}. We apply our adaptive context-aware object detection technique to two different state-of-the-art one-stage object detectors: YOLOv3 \cite{redmon2018yolov3}, and RetinaNet \cite{retiannet}.
In order to modify the static models into an adaptive context-aware network, we use a pre-trained backbone, and replace the detection head with our pool of specialized branches as well as our branch controller as explained in Section \ref{sec:clustering}.
We used Darknet-53 backbone with YOLOv3, and ResNet50 backbone with RetinaNet. 

\noindent \textbf{Hardware and Analysis Tools:} We deploy our object detection model on an NVIDIA Jetson Nano board as a representative device of modern embedded systems \cite{nano}. 
This board has a low-power embedded GPU, a Quad-core ARM Cortex CPU with 4 GB of RAM, and it uses about 5 to 10 Watts depending on the workload.
We analyzed the latency and the power consumption of our proposed technique on the Jetson Nano board.
We used Nvidia's Tegrastats utility to measure the latency and the power consumption.
We also used the COCO official APIs to evaluate the model accuracy. 

In this Section, we evaluate our spatial-context-based clustering by visualizing some of the formed clusters in Section \ref{context_based_clustering_evaluation}.
In Section \ref{branch_controller_evaluation}, we validate the ability of the branch controller to detect the right spatial context for the input images. 
After that, we analyze the overall performance and efficiency of our AdaCon models in Section \ref{adacon_general_sb_evaluation}.
Then, we show the effect of the introduced branch controller execution modes in Section \ref{bc_modes_eval}.
Finally, we analyze the accuracy-efficiency trade-off that our AdaCon models can achieve by choosing a different number of clusters, as well as different branch controller execution modes in Section \ref{adaptive_object_detection_evaluation}.

\begin{figure*}[t]
\centering
\begin{tabular}{c}
\subfloat[3 Clusters]{\includegraphics[trim=30 30 30 40, clip, width=0.33\textwidth,height=140pt,keepaspectratio]{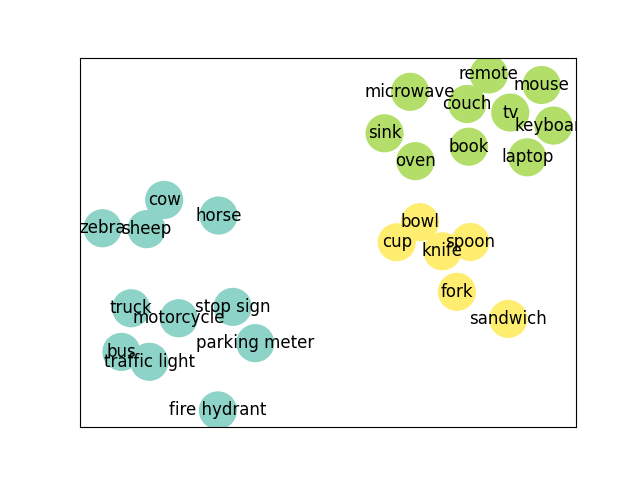}}
\subfloat[4 Clusters]{\includegraphics[trim=30 30 30 40, clip, width=0.33\textwidth,height=140pt,keepaspectratio]{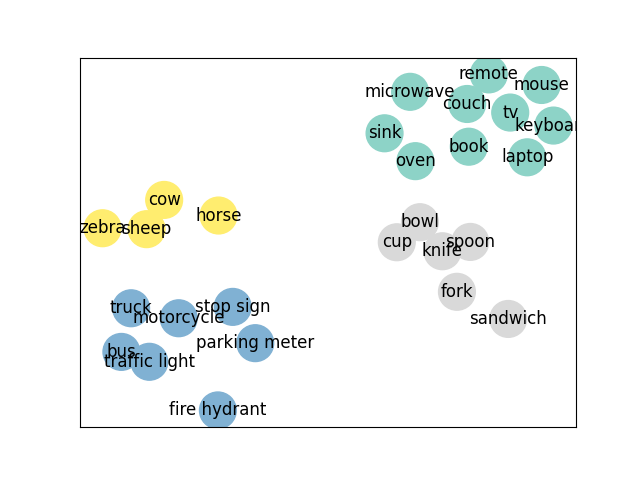}}
\subfloat[5 Clusters]{\includegraphics[trim=30 30 30 40, clip, width=0.33\textwidth,height=140pt,keepaspectratio]{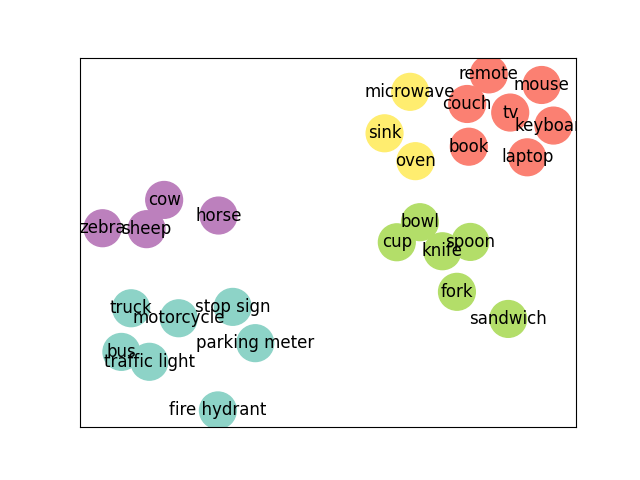}} 

\end{tabular}
\belowcaptionskip -10 pt
\caption{Our spatial-context-based clustering output for 30 object categories from the COCO dataset. The formed clusters become more fine-grained as the number of clusters increase.}
\label{clustering_evaluation}
\end{figure*}

\subsection{\textbf{Spatial-Context-Based Clustering Evaluation:}}
\label{context_based_clustering_evaluation}
To evaluate our spatial-context-based clustering technique, we visualize the formed clusters for a subset of the COCO dataset.
For the purpose of clear visualization only, we choose 30 representative object categories covering different spatial contexts (i.e., indoors, outdoors, street, farm, living room, kitchen, etc.). 
Figure \ref{clustering_evaluation} shows the result of our spatial-context-aware clustering technique when choosing different number of clusters.
Each node in Figure \ref{clustering_evaluation} represents a different object category, and the distance between any two objects represents the probability of their co-occurrence in the same scene (spatial context).
We can notice that as the number of clusters increases, the clustering automatically becomes more fine-grained. For example, clustering the objects into two clusters results in an indoor objects cluster, and another one for the outdoor objects.
Clustering the objects into four clusters, further categorizes the outdoor objects into objects that can be found on a farm, and others that can be found on a street, while categorizes the indoor objects to living room and kitchen objects. 

\vspace{-5pt}
\subsection{\textbf{Branch Controller Accuracy Evaluation:}}
\label{branch_controller_evaluation}
The branch controller takes the feature maps generated by the backbone for each image, and decides which branch(es) to execute based on the spatial context in the image (i.e., each spatial-context has a corresponding detection branch in our adaptive object detection model).
The ultimate goal of the branch controller is to determine the dominant spatial context in the input image.
The branch controller achieves that goal by assigning confidence scores that the input image belongs to each spatial context.
Then, the dominant context would be the one with the highest confidence score.

The branch controller accuracy is crucial for the accuracy of our adaptive model. The reason is that as the accuracy of the branch controller decreases, its ability to take the right decision about which branch(es) to execute decreases, and this directly affects the performance of the adaptive model.
To validate the decision-making capability of our branch controller, we analyze its accuracy in detecting the dominant spatial context on COCO dataset.
Table \ref{branch_controller_accuracy} shows the accuracy of the branch controller for different AdaCon models with different number of branches. 
As illustrated in the Table \ref{branch_controller_accuracy}, our branch controller is a light-weight model (i.e., the number of parameters and MAC operations are low) to prevent any memory, or compute overhead.
The results show that the branch controller accuracy is higher with fewer number of branches.
This is expected because its task becomes more complex as the number of branches increase.

\begin{center}

\begin{table}[t]
\caption{Branch Controller accuracy in detecting the correct spatial-context for different AdaCon models.}
\centering
\renewcommand{\arraystretch}{1.05}
\setlength{\tabcolsep}{2.5pt}

\begin{tabular}{ c | c | c | c | c | c} 
\toprule
\textbf{Model name} & \textbf{Backbone} & \textbf{Num.} & \textbf{Accuracy} & \textbf{Param} & \textbf{MAC}\\
& & \textbf{Branches} & \textbf{(\%)} &  \textbf{$(M)$} & \textbf{($Gops$)} \\
\midrule
Yolov3 & Darknet53 & 2 & 95.3 & \multirow{4}{*}{2.53} & \multirow{4}{*}{0.4} \\ 
Yolov3 & Darknet53 & 3 & 93.3 & &  \\ 
Yolov3 & Darknet53 & 4 & 93.2 &  &  \\ 
Yolov3 & Darknet53 & 5 & 91.7 &  &  \\ 
\midrule
\midrule
RetinaNet & Resnet50 & 2 & 93.3 & \multirow{3}{*}{0.55} & \multirow{3}{*}{0.17} \\
RetinaNet & Resnet50 & 3 & 90.9 & & \\
RetinaNet & Resnet50 & 4 & 89.5 & & \\

\bottomrule
\end{tabular}
\label{branch_controller_accuracy}
\vspace{-10pt}
\end{table}
\end{center}

\vspace{-25pt}

\subsection{\textbf{AdaCon Performance and Efficiency Evaluation:}}
\label{adacon_general_sb_evaluation}
\begin{figure*}[t]
\centering
\belowcaptionskip -5pt
\begin{tabular}{c}
\subfloat[\label{accuracy_adacon}]{\includegraphics[trim=10 10 10 0, clip, width=0.33\textwidth,keepaspectratio]{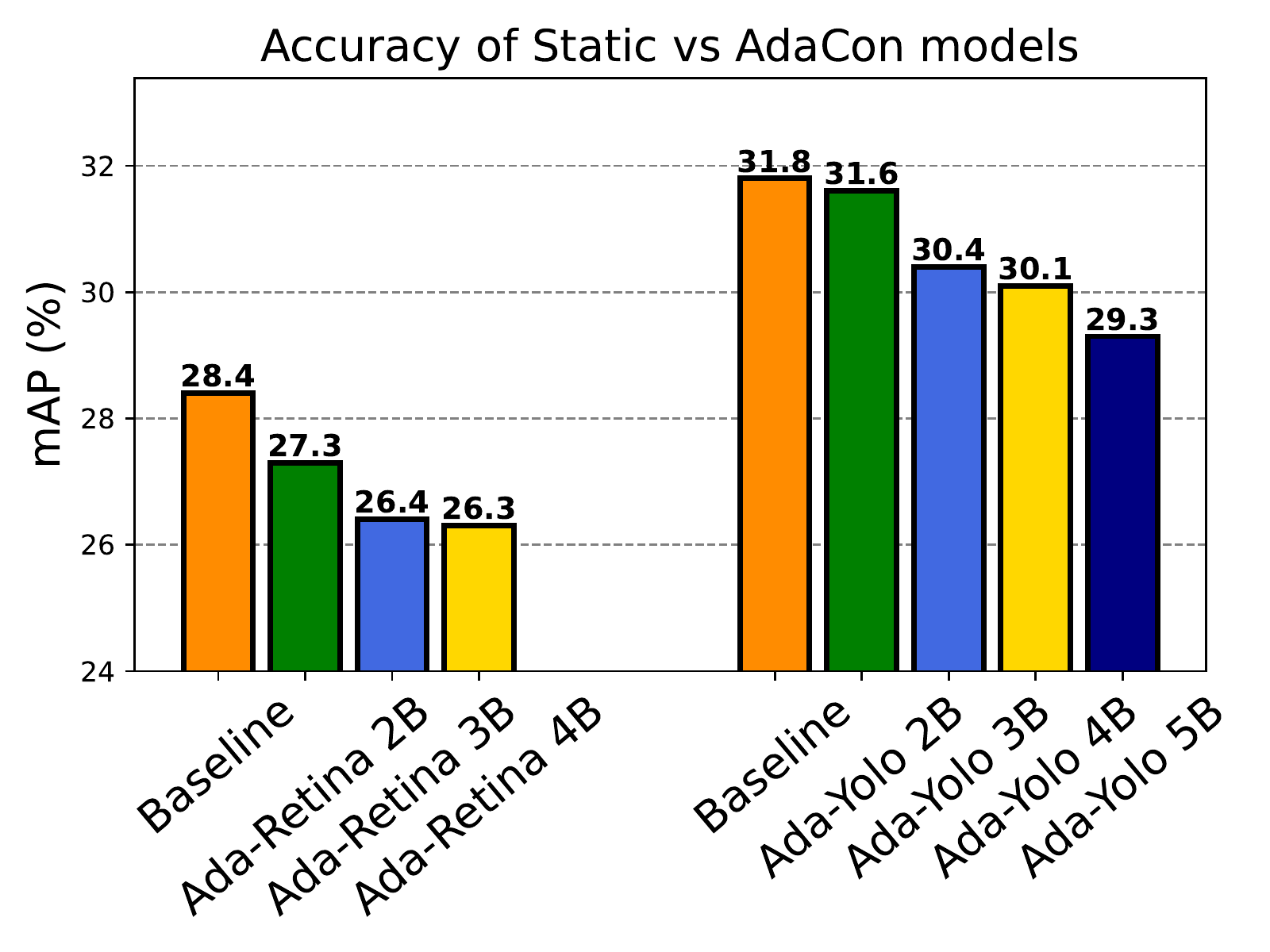}}
\subfloat[\label{latency_adacon}]{\includegraphics[trim=10 10 10 0, clip, width=0.33\textwidth,keepaspectratio]{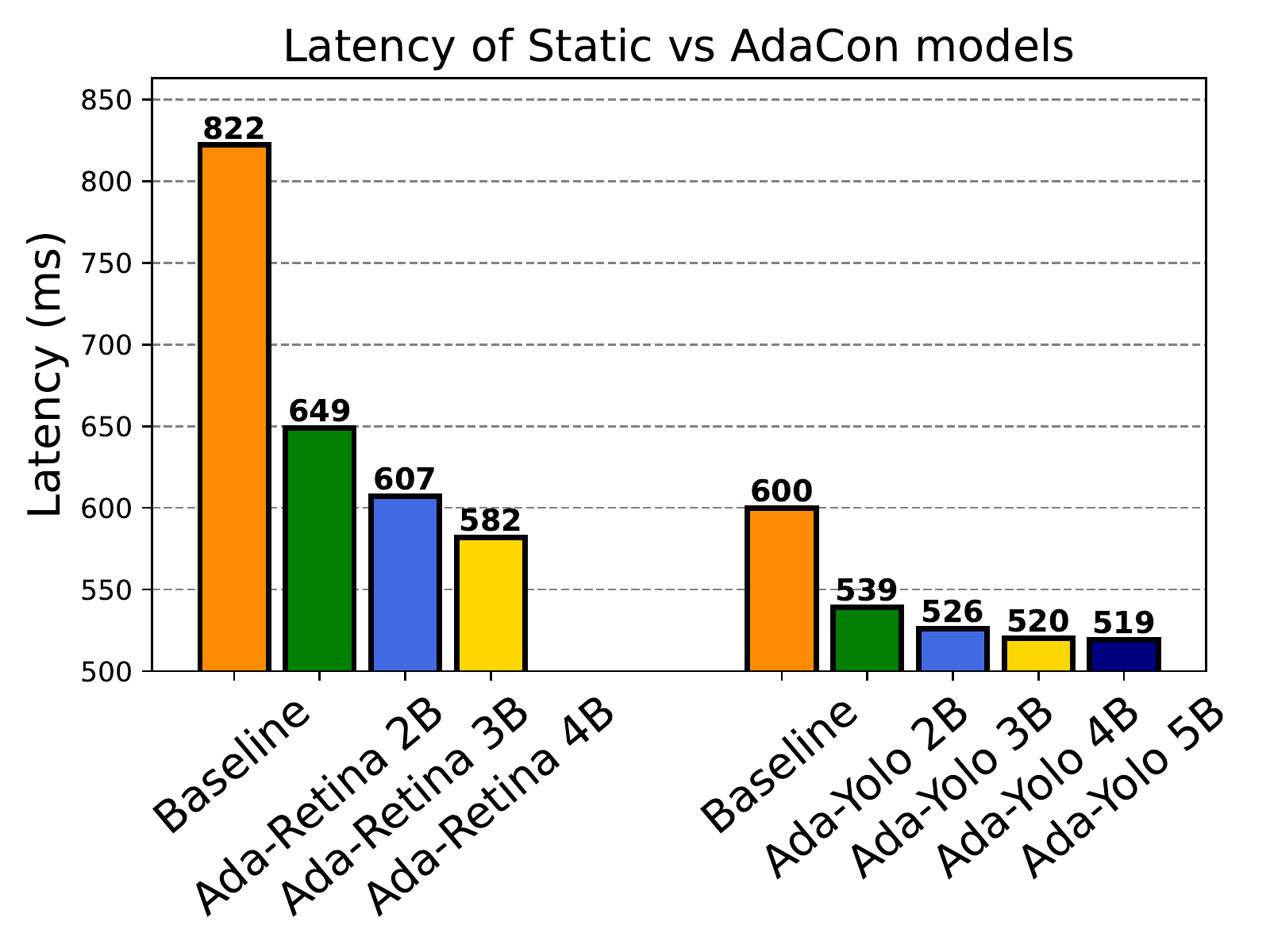}}
\subfloat[\label{energy_adacon}]{\includegraphics[trim=10 10 10 0, clip, width=0.33\textwidth,keepaspectratio]{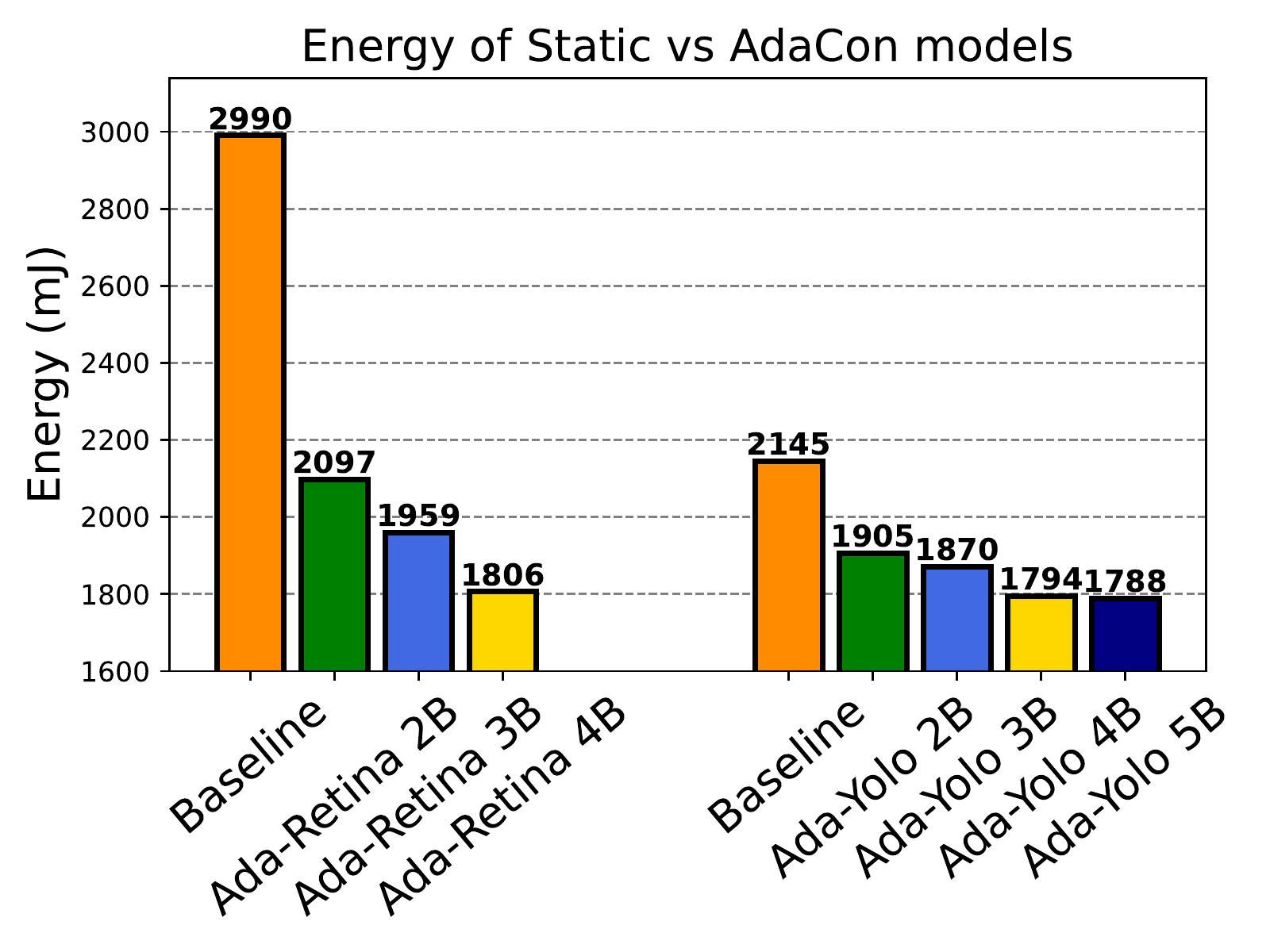}}

\end{tabular}
\caption{Evaluation of (a) Average Precision (b) Latency (c) Energy for AdaCon-RetinaNet and AdaCon-YOLO models with different number of branches under single-branch execution mode compared to the static baseline models. Input image resolution is $416 \times 416$.}
\label{branches_evaluation}
\vspace{-5pt}
\end{figure*}

To analyze the overall performance of our proposed adaptive technique, we show the average precision, the latency per inference, and the energy per inference of our AdaCon models with different number of branches compared to the static baselines in Figures \ref{accuracy_adacon}, \ref{latency_adacon}, and \ref{energy_adacon}, respectively.
The x-axis shows the different static and adaptive models with different number of branches (i.e., Ada-YOLO 2B denotes an AdaCon-YOLO model with two branches operating in the single-branch execution mode).
In Figure \ref{accuracy_adacon}, the y-axis shows the mean average precision of the object detection.
In Figures \ref{latency_adacon} and \ref{energy_adacon}, the y-axis represents the latency per inference in milliseconds, and the energy per inference in millijoules, respectively.
The results show that the average precision of the detection models decreases as the number of branches increases. The reason is that the accuracy of the branch controller decreases as the number of branches increases, leading to more misses by the branch controller. 
On the other hand, the energy and the latency decrease as the number of branches increase. The reason is that for models with more branches, each branch is responsible for a smaller subset of the object classes, hence it is more compact as explained in Section \ref{branch_architecture}.

\begin{figure*}[t]
\centering
\begin{tabular}{c}
\subfloat[Average Precision of AdaCon-YOLO \label{bc_yolo_accuracy}]{\includegraphics[trim=10 10 10 0, clip, width=0.33\textwidth,keepaspectratio]{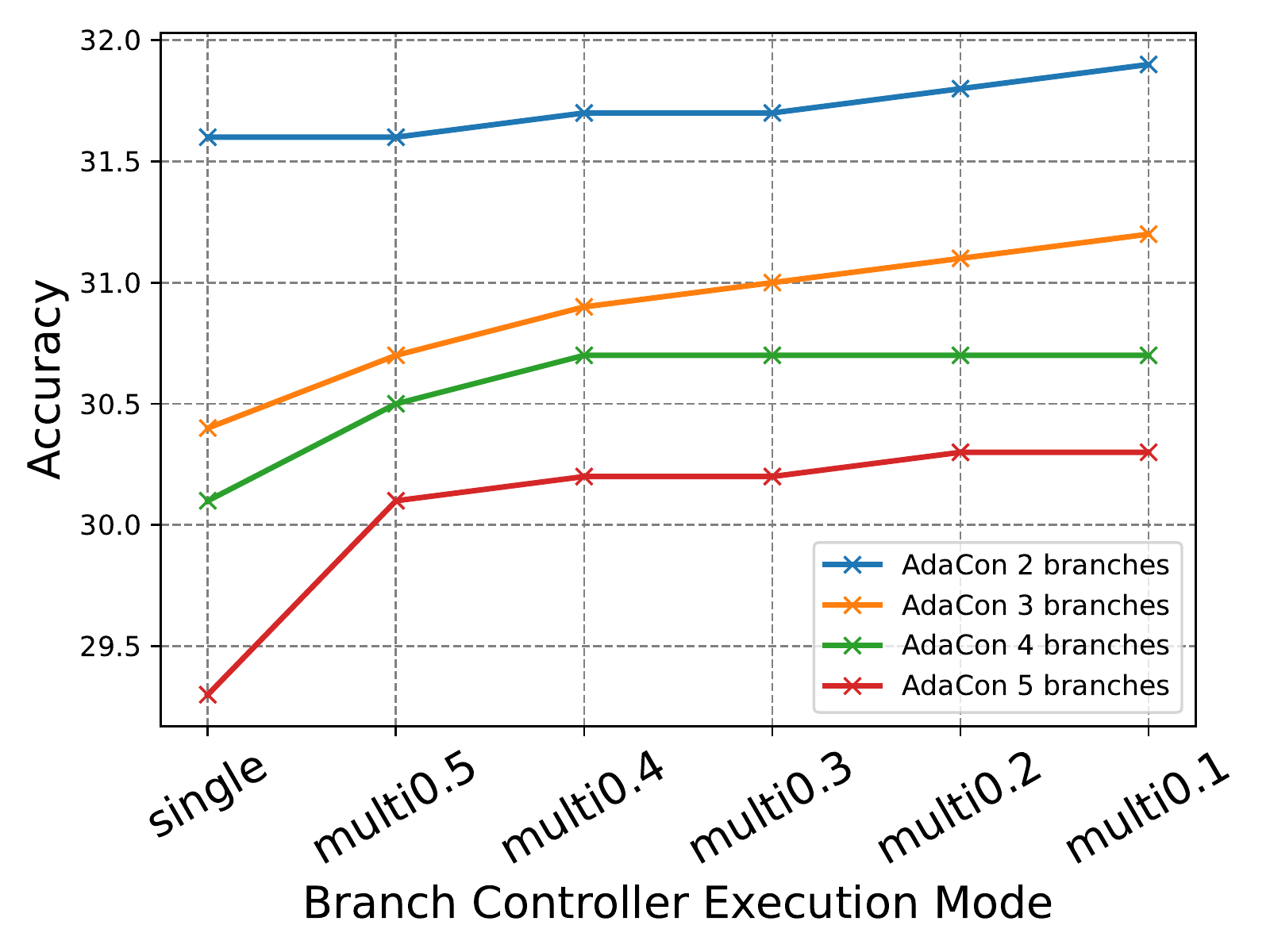}}

\subfloat[Latency of AdaCon-YOLO \label{bc_yolo_latency}]{\includegraphics[trim=10 10 10 0, clip, width=0.33\textwidth,keepaspectratio]{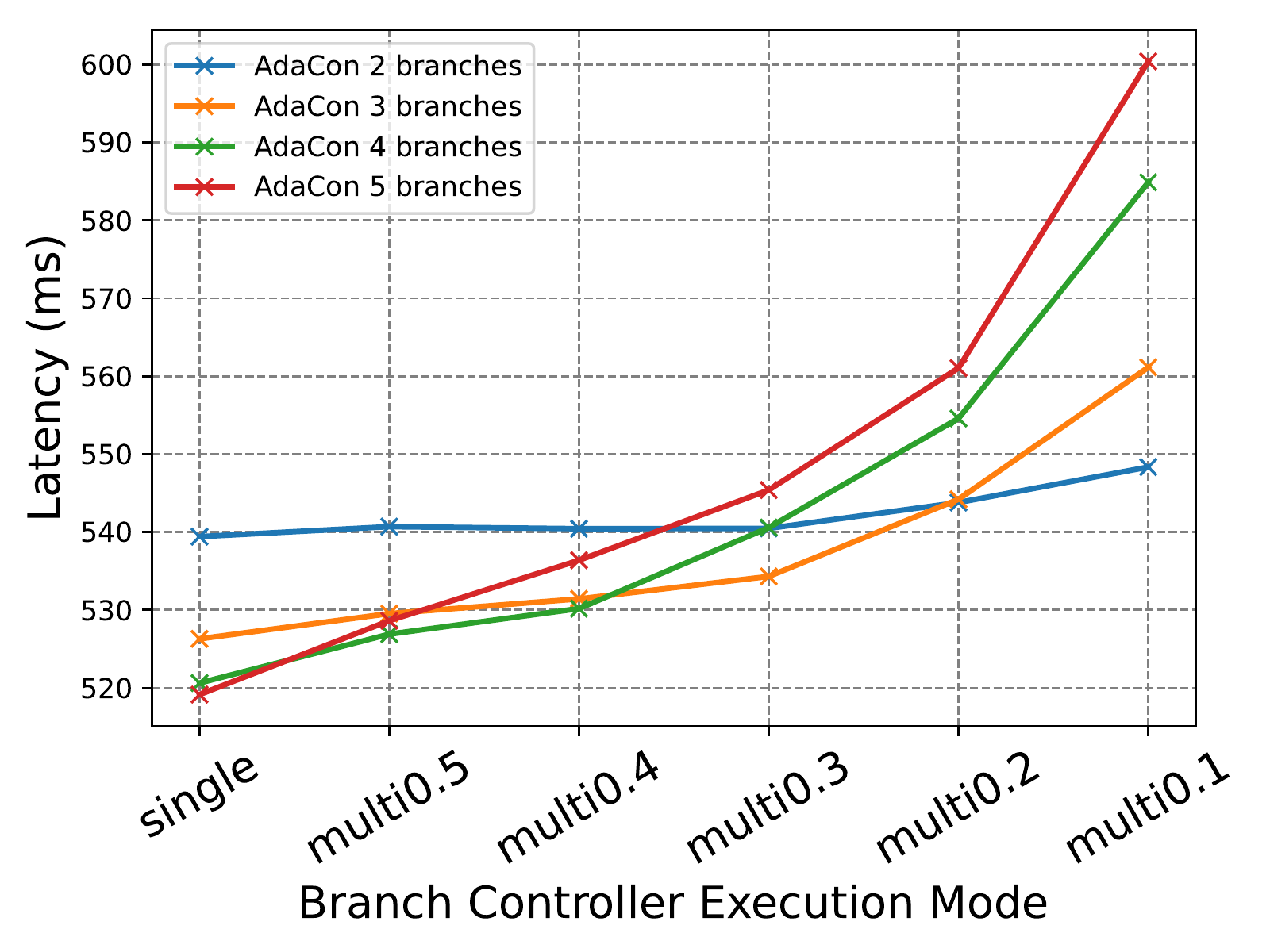}}

\subfloat[Energy of AdaCon-YOLO \label{bc_yolo_energy}]{\includegraphics[trim=10 10 10 0, clip, width=0.33\textwidth,keepaspectratio]{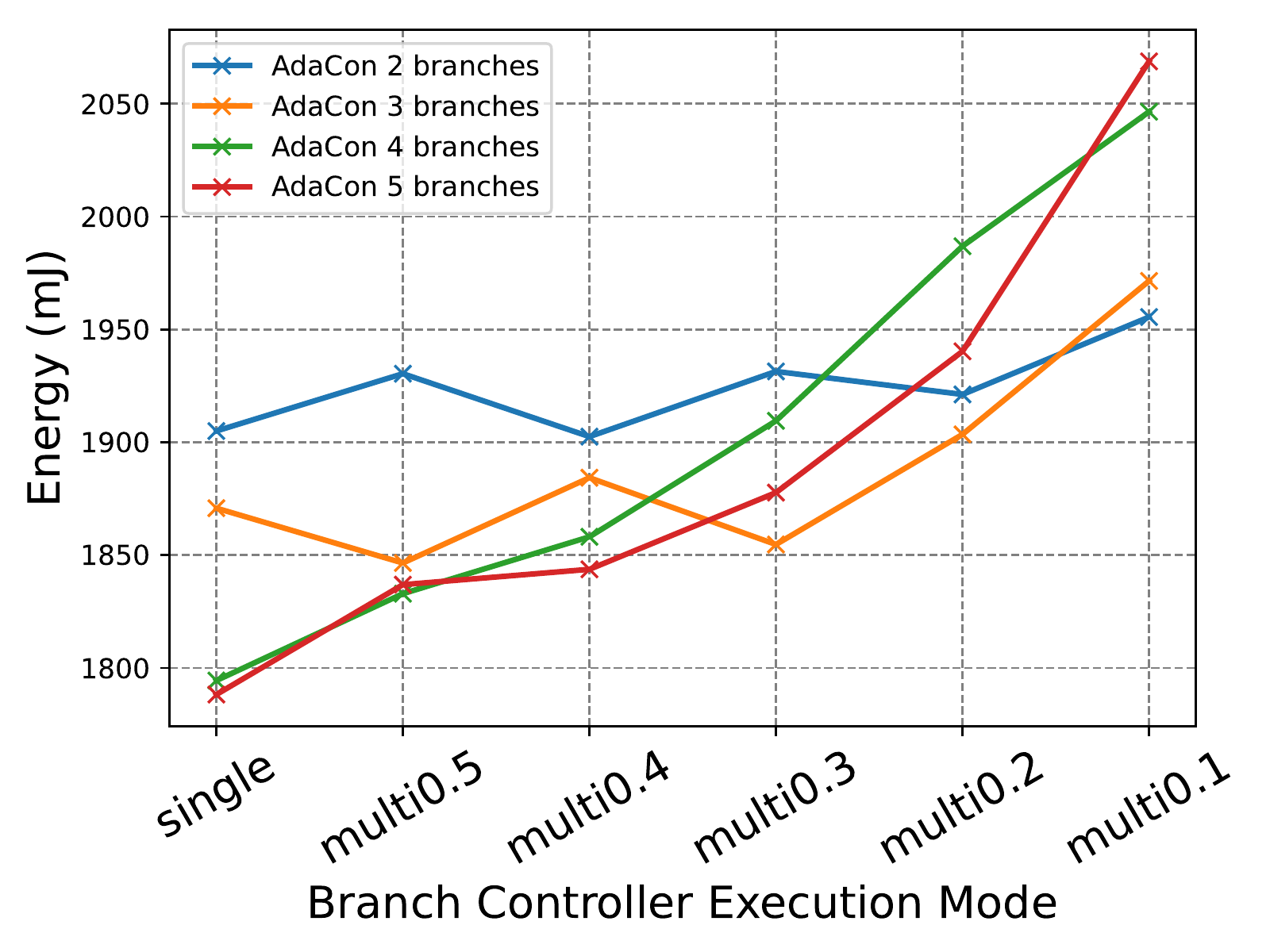}} \\

\subfloat[Average Precision of AdaCon-RetinaNet \label{bc_retina_accuracy}]{\includegraphics[trim=10 10 10 0, clip, width=0.33\textwidth,keepaspectratio]{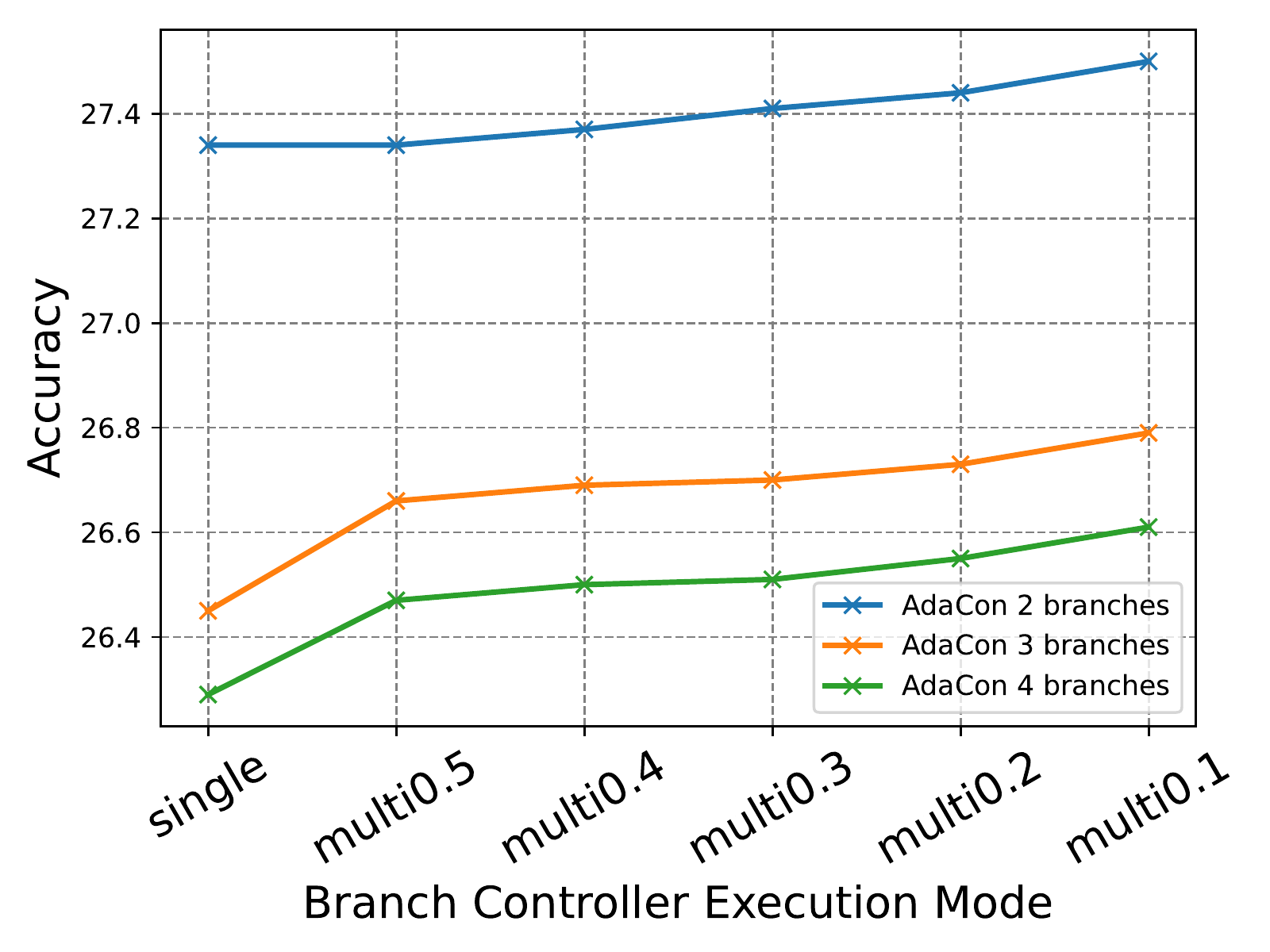}}

\subfloat[Latency of AdaCon-RetinaNet \label{bc_retina_latency}]{\includegraphics[trim=10 10 10 0, clip, width=0.33\textwidth,keepaspectratio]{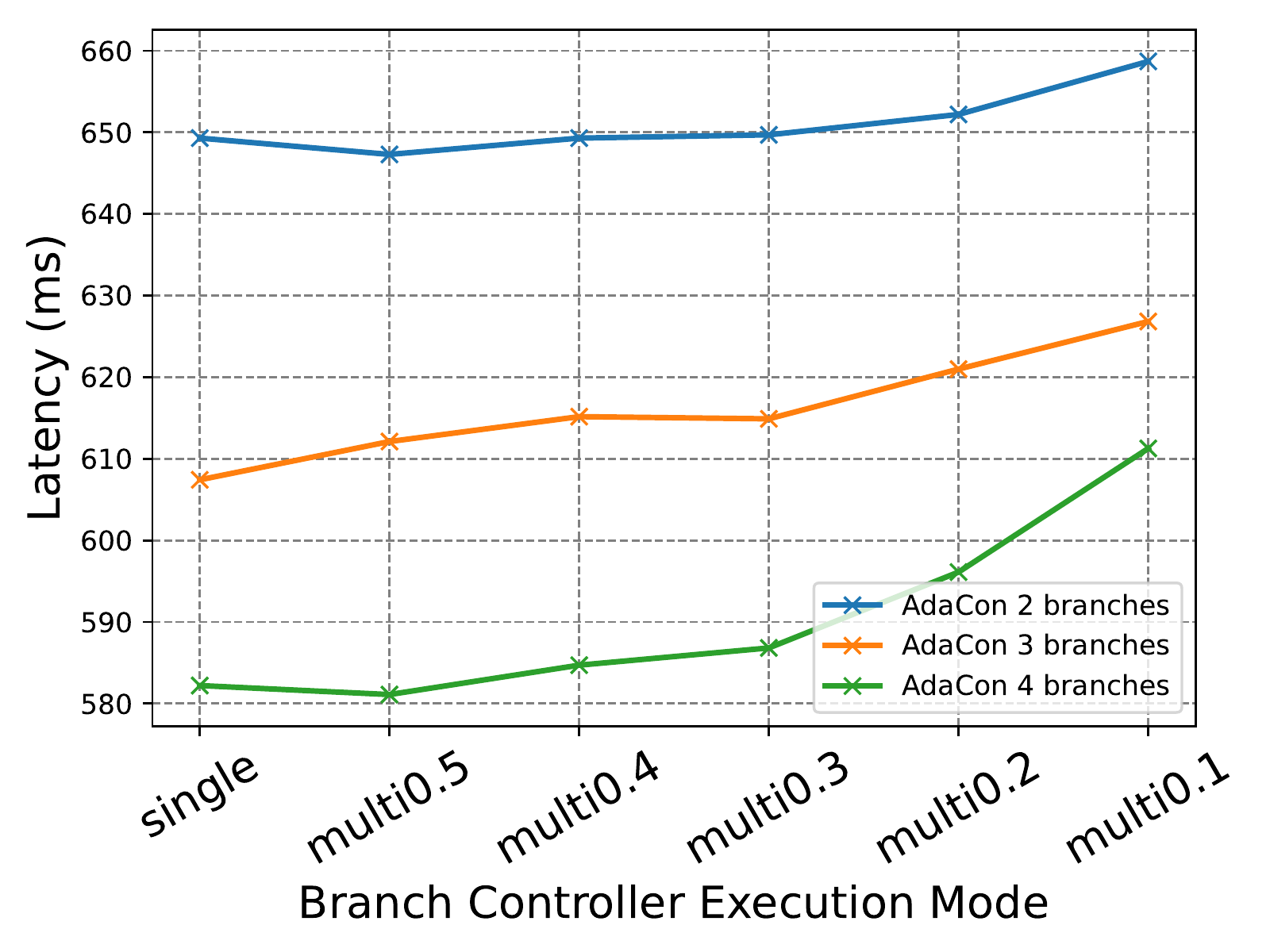}}

\subfloat[Energy of AdaCon-RetinaNet \label{bc_retina_energy}]{\includegraphics[trim=10 10 10 0, clip, width=0.33\textwidth,keepaspectratio]{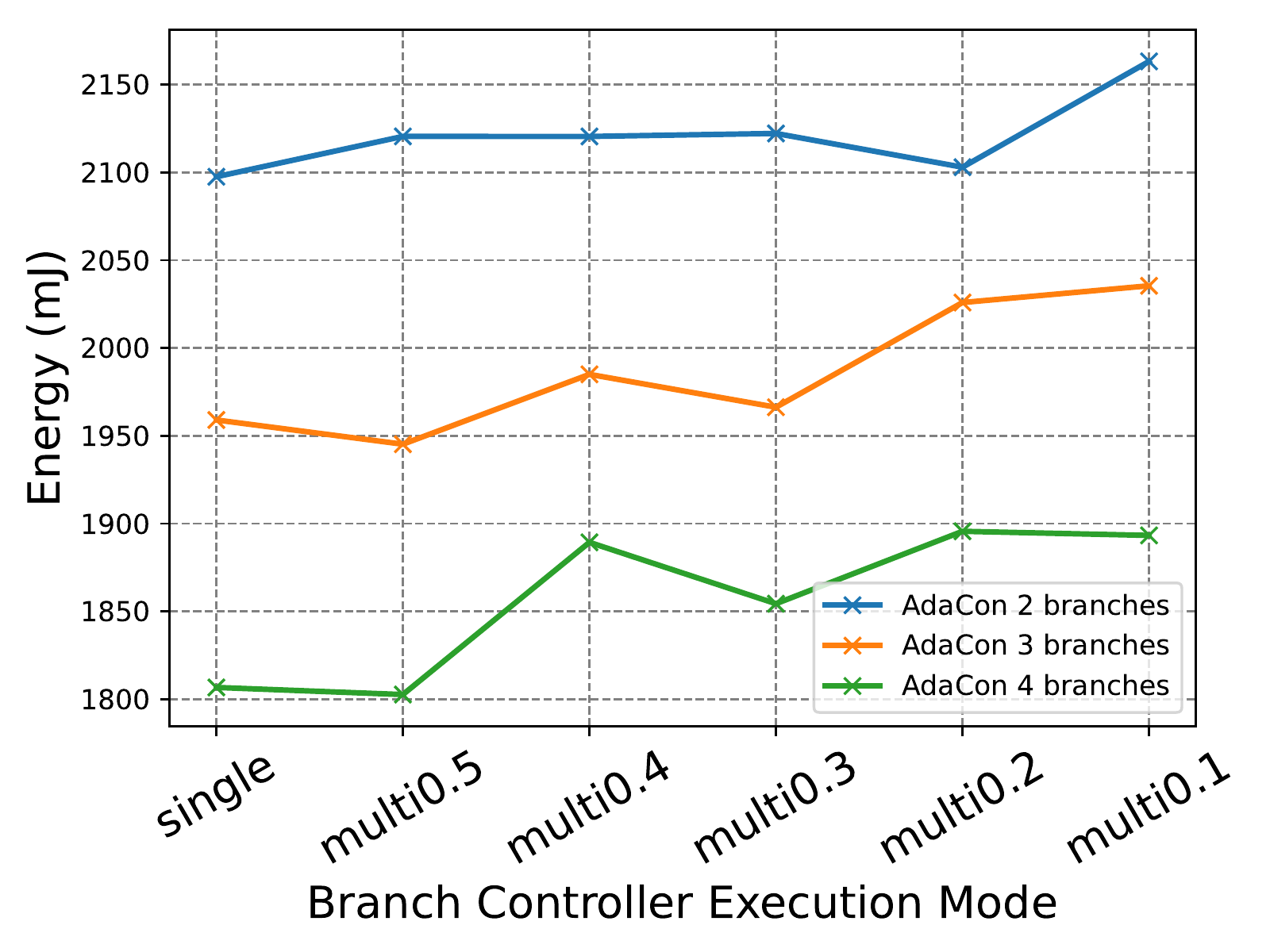}}

\end{tabular}
\caption{Evaluation of the Average Precision, the Latency and the Energy for AdaCon-YOLO and AdaCon-RetinaNet models under different branch controller execution modes. 
$single$ represents single-branch execution mode, while $multix$ represents multi-branch execution mode where $x$ represents the confidence score threshold used by the branch controller. Input image resolution is $416 \times 416$.}
\label{bc_eval}
\vspace{-15pt}
\end{figure*}

\begin{figure*}[t]
\centering

\begin{tabular}{c}
\subfloat[Energy/Accuracy trade-off for static and AdaCon-\\\hspace{\textwidth} RetinaNet models \label{pareto_retina_energy_416}]{\includegraphics[width=0.43\textwidth,keepaspectratio]{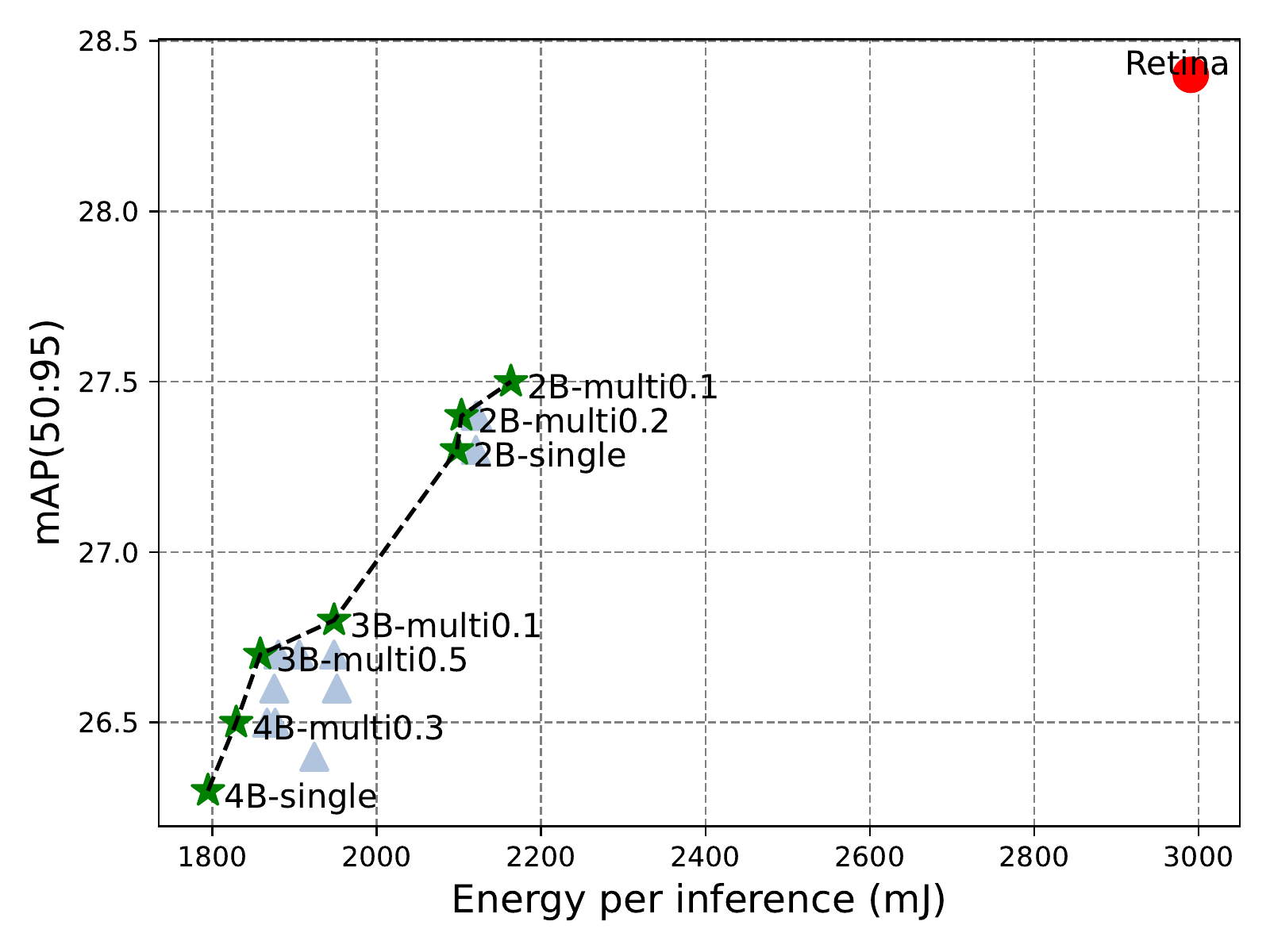}}

\subfloat[Latency/Accuracy trade-off for static and AdaCon-\\\hspace{\textwidth} RetinaNet models\label{pareto_retina_latency_416}]{\includegraphics[width=0.43\textwidth,keepaspectratio]{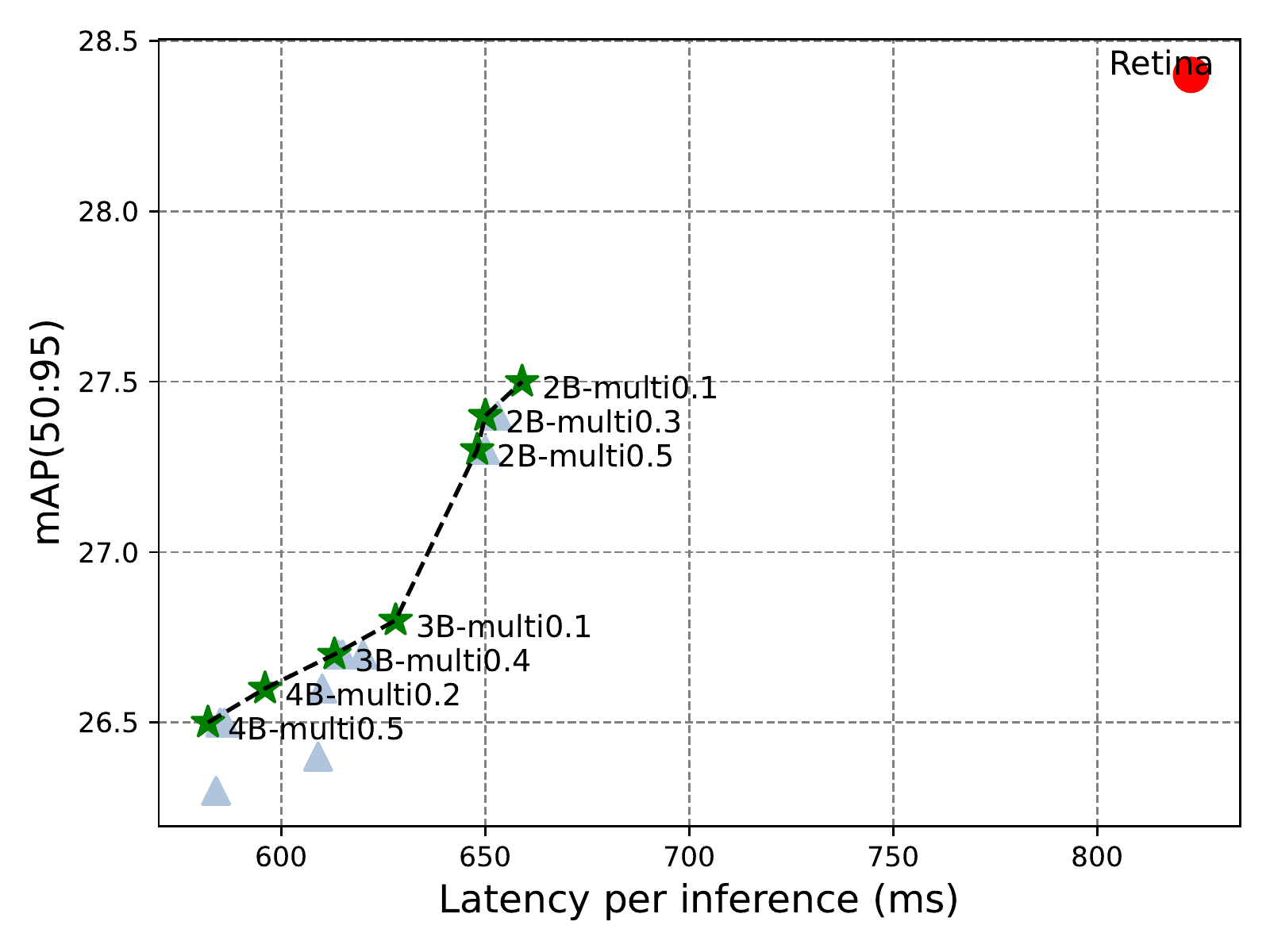}} \\

\subfloat[Energy/Accuracy trade-off for static and AdaCon-\\\hspace{\textwidth} YOLOv3 models \label{pareto_yolo_energy_416}]{\includegraphics[width=0.43\textwidth,keepaspectratio]{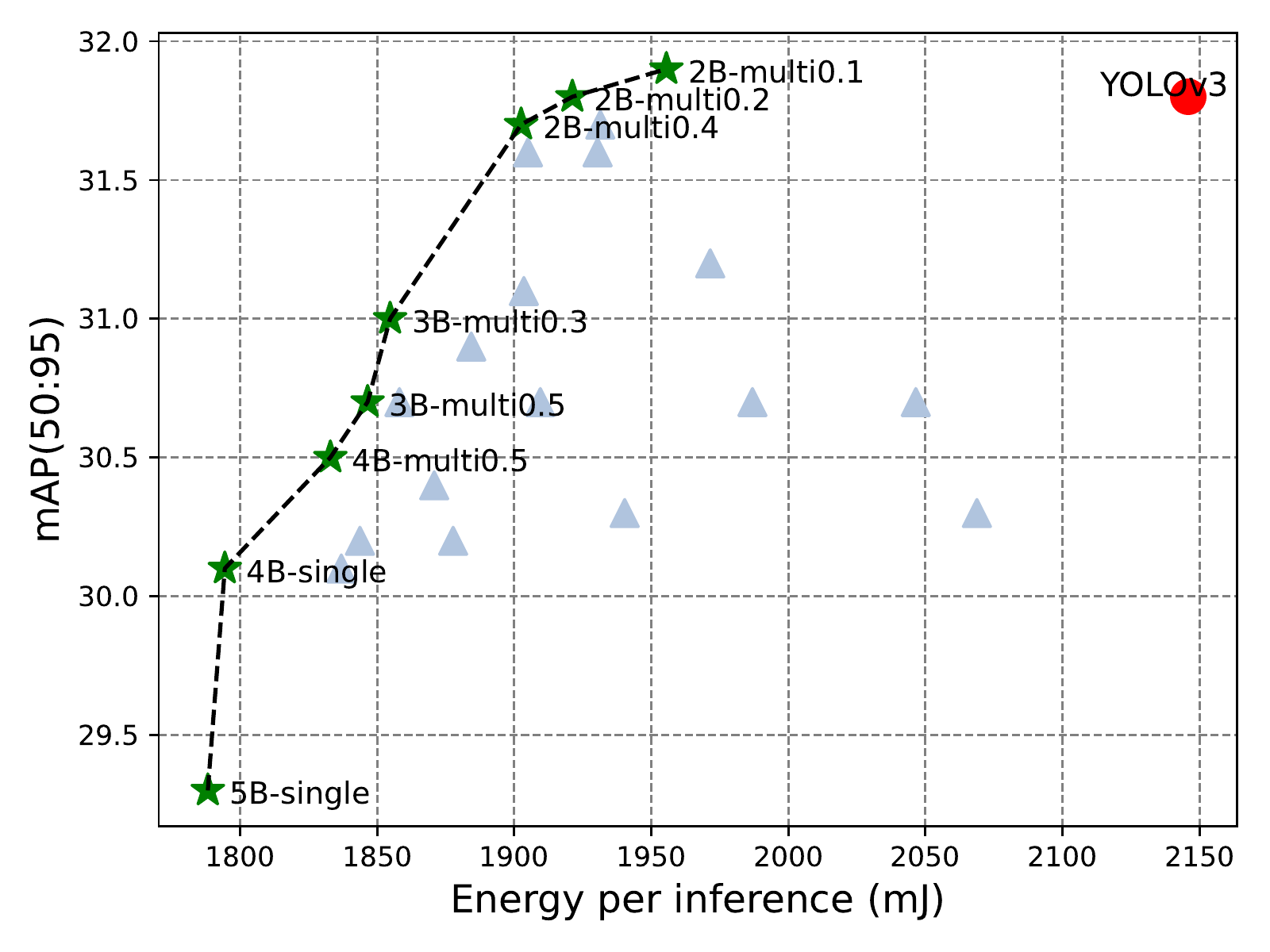}}

\subfloat[Latency/Accuracy trade-off for static and AdaCon- \\\hspace{\textwidth} YOLOv3 models\label{pareto_yolo_latency_416}]{\includegraphics[width=0.43\textwidth,keepaspectratio]{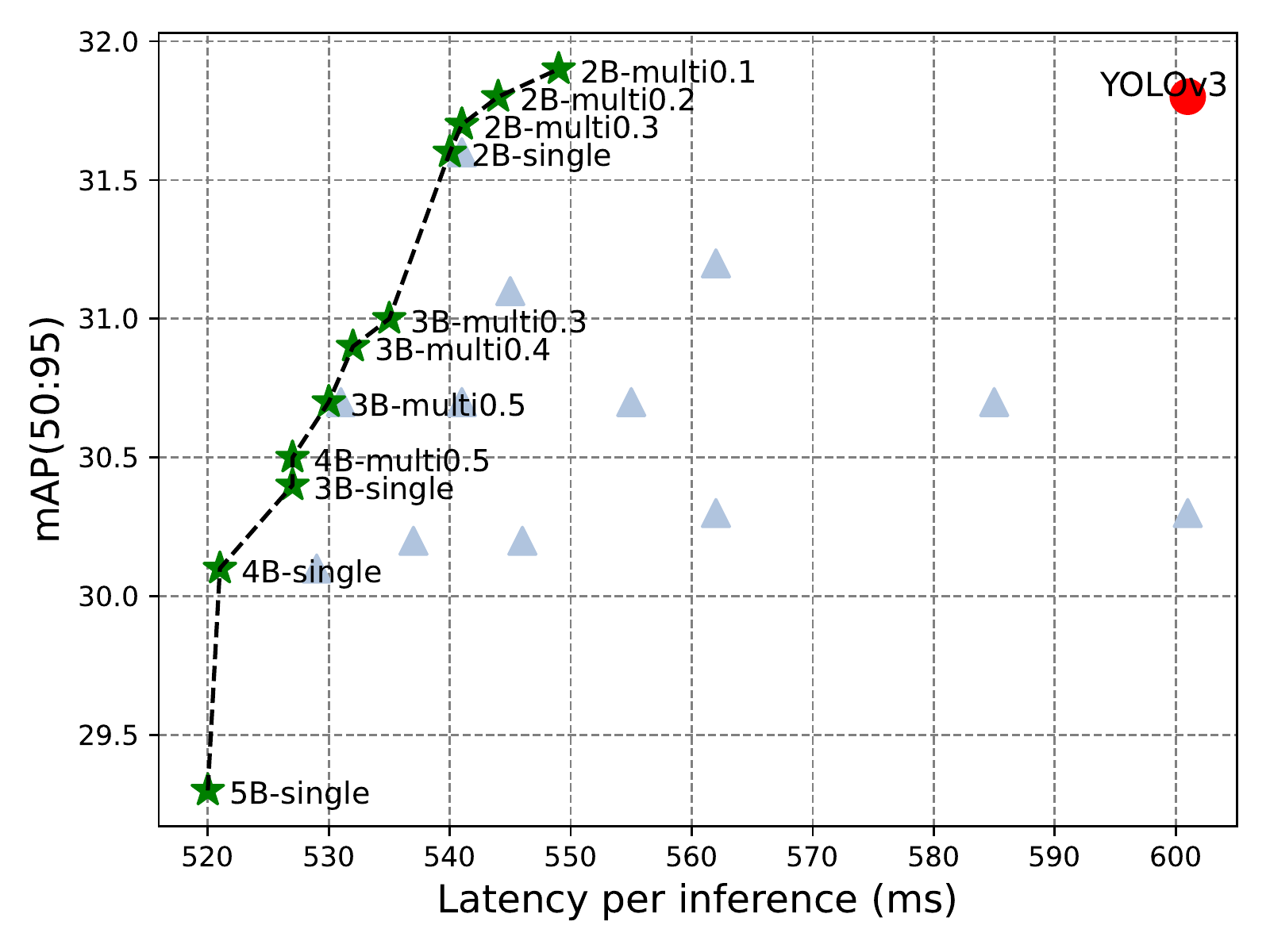}}

\end{tabular}
\caption{Energy, Accuracy and Latency Trade-offs for the different adaptive models. The adaptive models are named as (\textit{$n$}B-\textit{$mode$}), where \textit{$n$} refers to the number of branches in the AdaCon model, and \textit{$mode$} refers to the branch controller operation mode (i.e., \emph{single} or \emph{multi}), and the branch controller threshold in case of \emph{multi} execution.}
\label{pareto_evaluation_416}
\vspace{-10pt}
\end{figure*}

\begin{center}
\belowcaptionskip -10pt
\begin{table*}[t]
\small 
\caption{Results - AdaCon vs Static Models Accuracy and Efficiency Metrics - Latency and Energy are measured per inference. Sparam represents the total number of parameters for the model. Dparam represents the number of parameters used by the adaptive model at a time. MACs is the number of multiply-accumulate operations. Efficiency is $Accuracy/(Energy \times Latency)$}
\centering
\renewcommand{\arraystretch}{1.1}
\setlength{\tabcolsep}{1.2pt}
\begin{tabular}{ c | c | c c | c c | c c | c c | c c c | c c | c } 
\toprule
\textbf{Model} & \textbf{Image} & \multicolumn{4}{c}{\textbf{Accuracy}} & \multicolumn{2}{c}{\textbf{Latency}} & \multicolumn{2}{c}{\textbf{Energy}} & \textbf{Sparam} & \multicolumn{2}{c}{\textbf{Dparam}} & \multicolumn{2}{c}{\textbf{MACs}} & \textbf{Efficiency} \\
 & \textbf{Size}& $AP_{50:95}$ & \textbf{(\%)} & $AP_{50}$ & $AP_{75}$ & $(ms)$ & \textbf{(\%)} & $(mJ)$ & \textbf{(\%)}  & $(M)$ & $(M)$ & \textbf{(\%)} & $(G ops)$ & \textbf{(\%)} & \textbf{(\%)}\\
\midrule
\midrule
RetinaNet & 416 & 28.4 & - & 44.7 & 29.6 & 823.0 & - & 2990.4 & - & 32.44 & 0.00 & - & 41.08 & - & - \\

AdaCon 2B-multi 0.1 & 416 & 27.5 & \textbf{-3.2\%} & 43.5 & 28.6 & 659.0 & \textbf{-19.9\%} & 2163.2 & \textbf{-27.7\%} & 32.97 & 30.15 & \textbf{-7.1\%} & 26.50 & \textbf{-35.5\%} & \textbf{+67.2\%} \\

AdaCon 3B-multi 0.1 & 416 & 26.8 & \textbf{-5.6\%} & 42.8 & 27.8 & 628.0 & \textbf{-23.7\%} & 1947.9 & \textbf{-34.9\%} & 32.76 & 29.13 & \textbf{-10.2\%} & 20.98 & \textbf{-48.9\%} & \textbf{+89.9\%} \\

\midrule
\midrule
RetinaNet & 320 & 26.1 & - & 41.3 & 27.1 & 627.0 & - & 2349.7 & - & 32.44 & 0.00 & - & 24.27 & - & - \\
AdaCon 2B-multi 0.3 & 320 & 25.1 & \textbf{-3.8\%} & 40.1 & 26.1 & 504.0 & \textbf{-19.6\%} & 1521.5 & \textbf{-35.2\%} & 32.97 & 30.15 & \textbf{-7.1\%} & 15.49 & \textbf{-36.2\%} &  \textbf{+84.8\%} \\

AdaCon 3B-multi 0.5 & 320 & 24.4 & \textbf{-6.5\%} & 39.2 & 25.4 & 483.0 & \textbf{-23.0\%} & 1351.2 & \textbf{-42.5\%} & 32.76 & 29.13 & \textbf{-10.2\%} & 12.19 & \textbf{-49.8\%} & \textbf{+111.0\%} \\

AdaCon 4B-single & 320 & 24.0 & \textbf{-8.0\%} & 38.5 & 24.7 & 458.0 & \textbf{-27.0\%} & 1273.1 & \textbf{-45.8\%} & 32.97 & 28.74 & \textbf{-11.4\%} & 12.06 & \textbf{-50.3\%} & \textbf{+132.3\%} \\

\midrule
\midrule
YOLOv3 & 416 & 31.8 & - & 55.3 & 33.1 & 601.0 & - & 2145.9 & - & 61.92 & 61.92 & - & 33.01 & - & - \\
AdaCon 2B-multi0.1 & 416 & 31.9 & \textbf{+0.3\%} & 53.2 & 33.6 & 549.0 & \textbf{-8.7\%} & 1955.5 & \textbf{-8.9\%} & 58.31 & 51.53 & \textbf{-16.8\%} & 28.34 & \textbf{-14.1\%} & \textbf{+20.5\%} \\

AdaCon 3B-multi0.3 & 416 & 31.0 & \textbf{-2.5\%} & 52.1 & 32.6 & 535.0 & \textbf{-11.0\%} & 1854.7 & \textbf{-13.6\%} & 56.75 & 48.44 & \textbf{-21.8\%} & 27.11 & \textbf{-17.9\%} & \textbf{+26.7\%} \\

AdaCon 5B-single & 416 & 29.3 & \textbf{-7.9\%} & 49.7 & 30.5 & 520.0 & \textbf{-13.5\%} & 1788.2 & \textbf{-16.7\%} & 56.95 & 45.88 & \textbf{-25.9\%} & 26.10 & \textbf{-20.9\%} & \textbf{+27.8\%} \\

\midrule
\midrule
YOLOv3 & 320 & 29.3 & - & 52.0 & 30.4 & 414.0 & - & 1451.1 & - & 61.92 & 61.92 & - & 19.53 & - & - \\

AdaCon 2B-multi0.2 & 320 & 29.1 & \textbf{-0.7\%} & 49.1 & 30.6 & 380.0 & \textbf{-8.2\%} & 1256.3 & \textbf{-13.4\%} & 58.31 & 51.59 & \textbf{-16.7\%} & 16.78 & \textbf{-14.1\%} & \textbf{+25.0\%} \\

AdaCon 3B-multi0.5 & 320 & 28.0 & \textbf{-4.4\%} & 47.6 & 29.1 & 366.0 & \textbf{-11.6\%} & 1206.3 & \textbf{-16.9\%} & 56.75 & 47.88 & \textbf{-22.7\%} & 15.91 & \textbf{-18.5\%} & \textbf{+30.0\%} \\

AdaCon 4B-single & 320 & 27.3 & \textbf{-6.8\%} & 46.6 & 28.3 & 359.0 & \textbf{-13.3\%} & 1205.4 & \textbf{-16.9\%} & 56.90 & 46.56 & \textbf{-24.8\%} & 15.60 & \textbf{-20.1\%} & \textbf{+29.4\%} \\

\bottomrule
\end{tabular}
\label{results}
\vspace{-15pt}
\end{table*}
\end{center} 

\vspace{-25pt}
\subsection{\textbf{AdaCon Evaluation for the Branch Controller Execution Modes:}}
\label{bc_modes_eval}
As explained in Section \ref{sec:clustering}, our branch controller gives a confidence score for executing each branch based on the spatial context of the input image.
We introduce two modes of operation: \emph{single-branch execution ($single$)} mode where only the branch with the highest confidence score gets executed, and \emph{multi-branch execution mode ($multi$)}. 
In the \emph{multi-branch} execution, all the branches with a confidence score higher than a certain threshold are executed.
These execution modes not only boost the accuracy, but they also add some flexibility during runtime because they enable a trade-off between the efficiency and the accuracy without the need to use a different model, or even re-train the existing model.

To analyze the effects of the different branch controller execution modes, we compare the accuracy, energy, and latency of the adaptive models under different branch controller execution modes as shown in Figure \ref{bc_eval}.
The x-axis gives the different branch controller execution modes.
$single$ represents single-branch execution mode, while $multix$ represents multi-branch execution mode where $x$ represents the confidence score threshold used by the branch controller.
The y-axis shows the mean average precision, the latency per inference in milliseconds, and the energy consumption per inference in millijoules in Figures \ref{bc_yolo_accuracy}, \ref{bc_yolo_latency}, \ref{bc_yolo_energy}, respectively for AdaCon-YOLO, and in Figures \ref{bc_retina_accuracy}, \ref{bc_retina_latency}, and \ref{bc_retina_energy}, respectively for AdaCon-RetinaNet.
We can notice that as the branch controller threshold decreases, more branches are executed, boosting the accuracy. 
On the other hand, executing more branches adds an overhead to the latency and the energy of our adaptive model. 
Figures \ref{bc_yolo_latency}, and \ref{bc_yolo_energy} show that multi-branch execution for AdaCon models with a larger number of branches can have a bigger overhead on the latency and energy. 
This is reasonable because if the threshold for multi-branch execution is low, more branches get executed, adding a bigger computational overhead when compared to AdaCon models with a fewer number of branches.

\subsection{\textbf{Pareto-Frontier analysis for AdaCon:}}
\label{adaptive_object_detection_evaluation}
As mentioned in Section \ref{branch_controller_evaluation}, and \ref{bc_modes_eval}, AdaCon has two different knobs that can be tuned to achieve the required performance and efficiency trade-off.
These knobs are the number of branches, and the branch controller execution mode. 
We combine the settings from the two knobs, and generate 30 different adaptive models.
We name our adaptive models as (\textit{$nB$}-\textit{$mode$}), where \textit{$n$} refers to the number of branches in the AdaCon model, and \textit{$mode$} refers to the branch controller execution mode (i.e., $multi$ or $single$), along with the branch controller threshold in case of $multi$ execution.
In Figure \ref{pareto_evaluation_416}, we analyze the energy/accuracy and the latency/accuracy trade-offs for AdaCon-RetinaNet, and AdaCon-YOLO along with the static baselines. 
In this experiment, we use input resolution $416 \times 416$. 

Figures \ref{pareto_retina_energy_416}, and \ref{pareto_yolo_energy_416} show the energy/accuracy trade-off for AdaCon-RetinaNet, and AdaCon-YOLOv3, respectively.
The x-axis shows the energy per inference in millijoules, while the y-axis shows the average precision. 
The red circles represent the static baselines, and the green stars represent the Pareto-frontier adaptive models. 
The Pareto-frontier models maximize the accuracy, and minimize the energy. 
Finally, the dominated models (i.e., the models with lower accuracy, and higher energy consumption than other models) are shown as blue triangles.
Similarly, in Figures \ref{pareto_retina_latency_416}, and \ref{pareto_yolo_latency_416}, the green stars represent the pareto-frontier models, which maximize the accuracy, and minimize the latency. 
We can notice that the Pareto-frontier includes AdaCon models with a different number of branches, and different branch execution modes. 
The Pareto-frontier models cover a range of efficiency-performance trade-off, and provide run-time flexibility for our AdaCon models.

For RetinaNet, our AdaCon Pareto-frontier models achieve around 27\% to 45\% reduction in the energy consumption, and 20\% to 27\% reduction in latency, while losing less than two points of average precision.
Similarly, for YOLOv3, our adaptive Pareto-frontier models achieve around 8\% to 17\% reduction in the energy consumption, and 8\% to 13\% reduction in latency, while sometimes slightly increasing the accuracy.


To provide more insights on the performance of our adaptive object detection models, we show the detailed analysis for the static baseline, as well as some of the Pareto-frontier AdaCon models in Table \ref{results}. 
We analyze three different standard metrics for the model accuracy: $AP_{50}$, $AP_{75}$, and $AP_{50:95}$. $AP_{50}$ and $AP_{75}$ represent the average precision with intersection over union (IOU) $> 0.5$ and $> 0.75$, respectively. 
$AP_{50:95}$ is the mean of the average precision for IOU ranging from 0.5 to 0.95 with a step size of 0.05, so it is the most representative to the overall accuracy of the model.
We also analyze some efficiency metrics: the latency per inference in milliseconds, and energy per inference in millijoules. 
\emph{Static parameters (Sparam)} represents the total number of parameters for the model. 
\emph{Dynamic parameters (Dparam)} represents the number of parameters used by our adaptive model at a time.
We used \emph{Dparam} as an estimate of the memory footprint of our adaptive model. 
MACs is the number of multiply-accumulate operations measured in Giga operations. 

In general, we want to increase the accuracy, and reduce the energy as well as the latency of our models. 
That is why we define the \textit{Efficiency} metric as $Accuracy/(Energy \times Latency)$ to compare the overall quality of the static versus the adaptive models.
Our Pareto-frontier AdaCon models can achieve up to 1.32$\times$, and 30\% improvement in \textit{Efficiency} over the static baseline for RetinaNet, and YOLOv3, respectively. 
The results show that our AdaCon technique achieves higher efficiency for RetinaNet compared to YOLOv3.
The reason is that the reductions in latency and energy are directly proportional to the reduction in the number of MAC operations as shown in Table \ref{results}.
Our technique is mainly focused on the detection head component of the object detection model.
The detection head represents around 62\% and 25\% of the total number of MAC operations for RetinaNet and YOLOv3, respectively.
That is why the improvements are more significant for RetinaNet.

\vspace{-5pt}
\section{\textbf{Conclusion}}
\label{sec:conclusion}

In this paper, we present a novel spatial-context aware adaptive network for object detection. Using the prior knowledge about the co-occurrence of objects in the real world scenes, we categorize the objects according to their probability to occur jointly. Then, we design an energy-efficient adaptive model that consists of three parts: a backbone to extract the features in the images, a branch controller that route the output of the backbone to the right specialized branch(es) according to the spatial context of the input image, and a pool of context-specialized branches. Our method improves the overall model efficiency by up to 30\% for YOLOv3 and 132\% for RetinaNet. AdaCon reduces the energy consumption by 8\% to 45\%, the latency by 8\% to 27\%, with a small loss in the accuracy.

\bibliographystyle{IEEEtran}
\bibliography{refbib}

\end{document}